%% file: main.tex
\documentclass{applemlr}

\input{apple_preamble}

\usepackage{needspace}
\definecolor{gain}{HTML}{157F3D}
\definecolor{loss}{HTML}{B00020}
\providecommand{\added}[1]{#1}

\definecolor{revorange}{HTML}{D65F1E}

\newcounter{obscnt}
\newcounter{takecnt}
\newenvironment{observation}[1][]{%
  \par\refstepcounter{obscnt}%
  \noindent\textbf{Observation \theobscnt.}\space\ignorespaces}{\par}
\newenvironment{takeaway}[1][]{%
  \par\refstepcounter{takecnt}%
  \noindent\textbf{Takeaway \thetakecnt.}\space\ignorespaces}{\par}

\title{Beyond Visual CoT: Internalized Visual Thinking for Proactive Video Reasoning}

\author{Xiaoyu Zhu}
\author{Xinke Deng}
\author{Suresh Taddewadikar}
\author{Arnab Kumar Mondal}
\author{Zhongyu Jiang}
\author{Ian Fasel}
\author{Joerg Liebelt}

\affiliation{Apple}

\abstract{
Multimodal large language models increasingly use visual chain-of-thought (Visual CoT) to reason about spatial, temporal, and embodied environments. By generating intermediate reasoning images, Visual CoT provides an intuitive mechanism for visual foresight but introduces substantial inference overhead, which is particularly problematic for proactive video reasoning. We ask whether models can learn to think visually during training while reasoning directly at inference. We introduce Internalized Visual Thinking (IVT), a post-training framework that jointly optimizes textual prediction and next-embedding prediction over unlabeled videos. Given a partially observed video, IVT predicts latent representations of future frames together with the target textual answer, encouraging the model to capture motion, object transitions, interactions, and latent intent. At inference, IVT generates the answer directly without synthesizing or re-encoding future frames. We conduct controlled studies across target representations, decoder designs, prediction horizons, data mixtures, training curricula, and predictive objectives. IVT improves over text-only post-training across all six evaluation settings while retaining the same efficient inference pathway. Compared with Visual CoT, IVT achieves comparable or better performance and reduces end-to-end latency by more than $5\times$. Our findings suggest that explicit pixel-space generation at inference time, as used in Visual CoT, may not be necessary for proactive video reasoning. Predictive world modeling can be internalized during training to produce multimodal reasoners that are both more accurate and substantially more efficient.
}

\correspondence{\sffamily Xiaoyu Zhu: \url{xiaoyu_zhu2@apple.com}}
\date{\sffamily\today}

\begin{document}

\maketitle

\input{sections/1_intro}
\input{sections/3_method}
\input{sections/4_experiment}
\input{sections/2_related}

\input{sections/5_conclusion}

\bibliographystyle{plainnat}
\bibliography{main}

\input{sections/appendix}

\end{document}

%% file: apple_preamble.tex
\usepackage{amsmath}
\usepackage{enumerate}
\usepackage{algorithm}
\usepackage{algpseudocode}
\usepackage{amsfonts}
\usepackage{amsthm}
\usepackage{cleveref}
\usepackage{diagbox}
\usepackage{colortbl}
\usepackage{amssymb}
\usepackage{xspace}
\usepackage{wrapfig}
\usepackage{adjustbox}
\usepackage{tabularx}
\usepackage{booktabs}
\usepackage{mathtools}
\usepackage{tikz}
\usepackage{enumitem}
\usepackage{silence}
\usepackage{dsfont}
\usepackage[table]{xcolor}
\usepackage[dvipsnames]{xcolor}
\usepackage{multirow}
\usepackage{makecell}
\usepackage{xfakebold}

\definecolor{textgray}{HTML}{6E6E73}
\usetikzlibrary{positioning, calc}
\usetikzlibrary{decorations.pathmorphing}

\makeatletter
\patchcmd{\wrong@fontshape}{\@gobbletwo}{}{}{}
\makeatother
\numberwithin{equation}{section}
\makeatletter
\AtBeginDocument{
  \urlstyle{sf}
  
}
\makeatother

\definecolor{light}{RGB}{125, 125, 125}
\crefname{tcb@cnt@pbox}{code}{code}
\Crefname{tcb@cnt@pbox}{Code}{Code}
\crefname{assumption}{assumption}{assumption}
\Crefname{assumption}{Assumption}{Assumptions}

\newtcolorbox[auto counter]{pbox}[2][]{
  colback=white,
  title=Code~\thetcbcounter: #2,
  #1,fonttitle=\sffamily,
  fontupper=\sffamily,
  arc=2pt,
  colframe=bgcolor,
  coltitle=fgcolor,
  colbacktitle=bgcolor,
  toptitle=0.25cm,
  bottomtitle=0.125cm
}

\makeatletter
\newcommand\applefootnote[1]{%
  \begingroup
  \renewcommand\thefootnote{}%
  \renewcommand\@makefntext[1]{\noindent##1}%
  \footnote{#1}%
  \addtocounter{footnote}{-1}%
  \endgroup
}
\makeatother

\definecolor{cverbbg}{gray}{0.90}

%% file: sections/1_intro.tex
\section{Introduction}

Intelligent systems operating in the visual world must do more than recognize what is currently visible. They must infer how an unfolding event will evolve and anticipate what is likely to happen next. Chain-of-thought (CoT) has substantially improved the reasoning ability of multimodal large language models (MLLMs) to solve problems by decomposing them into intermediate steps before producing an answer~\citep{wei2022chain,zhang2023multimodal}. Most existing MLLMs instantiate this process as \emph{textual CoT}, in which intermediate computation is represented as a sequence of natural-language tokens. Although effective, language is an imperfect reasoning representation for visual dynamics. Textual rationales can be unnecessarily long or redundant, increasing autoregressive decoding cost without consistently improving accuracy~\citep{wu2025more,xu2025chain}. More fundamentally, generated rationales are not guaranteed to faithfully reflect the computation that determines the final prediction~\citep{zhaorobustness,turpinlanguage,lanham2023measuring}. In multimodal settings, this problem is compounded by a modality mismatch: text-based reasoning can rely excessively on linguistic priors, hallucinate visual details, or fail to remain grounded in the spatial and temporal evidence contained in the input~\citep{wu2026grounded,kancheti2026chain}.

These limitations have motivated a growing shift from \emph{reasoning in words} toward \emph{reasoning in images}. Existing approaches construct visual reasoning by extracting relevant regions, annotating or transforming the input, drawing intermediate sketches, or generating new images as part of the reasoning process~\citep{zhou2024image,hu2024visual,chern2025thinking}. Such visual chains of thought can preserve geometry, object states, motion, and other information that is difficult to express precisely in language. In temporal and embodied settings, a particularly compelling formulation is \emph{predictive Visual CoT}: the model first generates an image of a plausible future or goal state and subsequently conditions its answer or action on that visual prediction~\citep{zhao2025cot,li2025zebra}. By externalizing the model's thinking of the future, predictive Visual CoT provides an intuitive mechanism for visual foresight and has demonstrated strong results on spatial reasoning, planning, and robotic manipulation.

However, explicitly rendering visual thoughts transfers a substantial computational burden to inference.
Unlike textual reasoning, which autoregressively decodes discrete tokens, Visual CoT requires decoding dense visual representations into an intermediate image and subsequently re-encoding that image into visual tokens for downstream reasoning, resulting in substantially higher computational cost.
This inference cost is particularly problematic for latency-critical applications in proactive reasoning, where a prediction loses value if it arrives after the event has already occurred.

\begin{figure}[t]
\centering
\includegraphics[width=\linewidth]{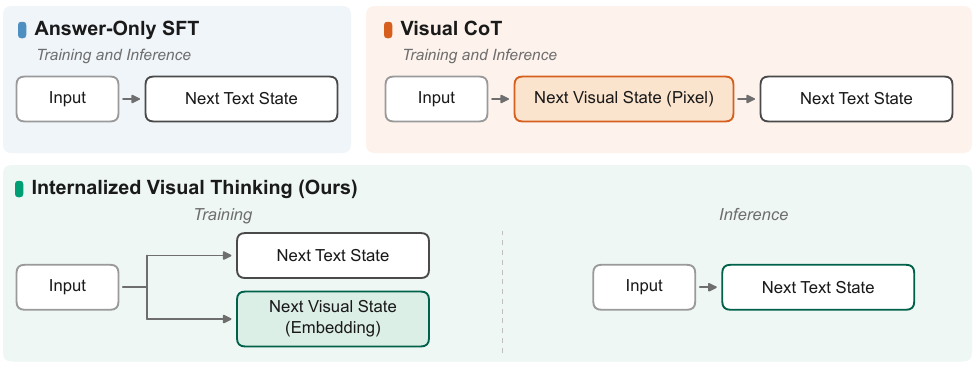}
\caption{\textbf{Three post-training paradigms for proactive video reasoning.} \emph{Answer-Only SFT} maps the observed input directly to the predicted textual state. \emph{Visual CoT} first generates an explicit future visual state in \emph{pixel space} and then conditions on it to produce the answer, so it pays for the image generation cost at inference. In contrast, our \emph{Internalized Visual Thinking} (IVT) predicts the future visual state as a latent \emph{embedding} target \emph{only during training}; at inference it generates the answer directly, without synthesizing an intermediate image, and therefore executes the same computation graph as Answer-Only SFT.}
\label{fig:teaser}
\vspace{-3mm}
\end{figure}
Following the definition in \citet{zhao2021review}, we consider two primary tasks for proactive video reasoning. \emph{Early-event prediction} requires recognizing an ongoing action before it is completed, whereas \emph{next-event prediction} requires anticipating an action that has not yet begun and may occur several seconds into the future~\citep{zhao2021review,sadegh2017encouraging}. Unlike offline video question answering, these tasks cannot rely on evidence that is already fully observable. Instead, models must infer latent intent, causal dynamics, and plausible future states from incomplete observations in a timely manner.
This motivates our central question:

\begin{quote}
\centering
\emph{Can MLLMs learn to think visually during training, yet reason efficiently at inference without explicitly generating their visual thoughts?}
\end{quote}

We answer this question through Internalized Visual Thinking (IVT), a post-training framework that learns predictive world representations from unlabeled videos. Rather than generating a future image at inference time, IVT jointly trains the model to (i) predict the latent embeddings of future frames from the observed video and (ii) generate the corresponding textual prediction, as shown in Figure~\ref{fig:teaser}. The future embeddings provide dense supervision that requires the model to capture object transitions, motion, interactions, and scene dynamics.
After post-training, the model directly produces an answer from the observed video. In this sense, IVT reformulates visual foresight from an explicit inference-time operation into the model's internal representations.

The formulation's effectiveness depends on several nontrivial design choices.
We conduct controlled experiments across multiple model architectures, target representations, prediction horizons, multimodal data-mixture ratios, loss functions, batching strategies, and curriculum designs.
Beyond reporting the best configuration, we systematically analyze both successful and unsuccessful training recipes, which demonstrates when future prediction improves reasoning and when it instead interferes with language generation.

Our experiments yield three principal findings. First, next-embedding prediction can substantially improve proactive video reasoning over text-only post-training. This result is notable because recent studies of unified generation and understanding suggest that adding generative capabilities to MLLMs does not consistently improve and can sometimes degrade the visual question answering performance~\citep{wen2026unig2u,cheng2026video}. Our results demonstrate that the value of generation lies not necessarily in synthesizing high-quality images, but in learning predictive representations that benefit proactive reasoning tasks.

Second, internalized visual thinking is not merely a cheaper approximation of explicit Visual CoT. Under matched base models, training data, and evaluation protocols, IVT surpasses strong Visual CoT counterparts on 4 of 6 benchmarks. These results indicate that Visual CoT may not be necessary for the proactive video reasoning task. Predictive world modeling can be internalized during training to produce efficient multimodal reasoners.

Third, the gains do not emerge automatically from adding an auxiliary prediction loss. We find that sharing the language-model decoder, carefully controlling the ratio between visual-prediction and text-generation for data batching, selecting an appropriate predictive horizon, and introducing the objectives through a suitable curriculum are all critical. These findings turn IVT from an isolated model improvement into a set of practical principles for building efficient MLLMs.

Together, our results establish a training-time alternative to Visual CoT: rather than asking a model to \emph{render} what it imagines, we teach it to encode predictive visual knowledge in the representations used to reason. This produces MLLMs that retain the foresight benefits of visual world modeling while preserving the inference efficiency required by proactive reasoning tasks.

%% file: sections/3_method.tex
\section{Internalized Visual Thinking}
\label{sec:method}

\subsection{Method Overview}

Let $X_{\leq t}=(I_1,\ldots,I_t)$ denote the observed video prefix up to the observation boundary $t$, let $q$ denote the textual prompt, and let $y=(y_1,\ldots,y_T)$ denote the target textual response. A standard supervised fine-tuning (SFT) objective minimizes the token-averaged negative log-likelihood
\begin{equation}
\mathcal{L}_{\mathrm{text}}
=
-\frac{1}{T}
\sum_{j=1}^{T}
\log p_{\theta}
\left(
y_j
\mid
y_{<j},X_{\leq t},q
\right),
\label{eq:text_loss}
\end{equation}
where $\theta$ denotes the trainable MLLM parameters. Although this objective teaches the model to map partial observations to a target response, it does not explicitly constrain the learned representation to capture how the visual state may evolve beyond the observation boundary.

IVT augments language supervision with an auxiliary objective that predicts latent representations of future video frames. Let $\mathcal{H}$ denote a set of future temporal offsets and let $K=|\mathcal{H}|$. For each $h\in\mathcal{H}$, the corresponding future frame is $I_{t+h}$. A target encoder $E_{\mathrm{tar}}$ maps this frame to a sequence of $M$ latent tokens,
\begin{equation}
Z_h
=
E_{\mathrm{tar}}(I_{t+h})
=
(z_{h,1},\ldots,z_{h,M})
\in \mathbb{R}^{M\times d},
\label{eq:target_latent}
\end{equation}
where $d$ denotes the target-feature dimension.
Conditioned only on the observed context $(X_{\leq t},q)$, the MLLM produces predictive hidden states at designated future-prediction positions. A projection head $P_{\phi}$ maps these hidden states into the target representation space,
\begin{equation}
\widehat{Z}_h
=
P_{\phi}
\left(
F_{\theta}(X_{\leq t},q;h)
\right)
\in \mathbb{R}^{M\times d},
\label{eq:predicted_latent}
\end{equation}
where $F_{\theta}(\cdot;h)$ denotes the predictive hidden states associated with future offset $h$, and $\phi$ denotes the projection-head parameters.
We optimize the predicted future representations using a predictive objective $\mathcal{L}_{\mathrm{pred}}$. The overall IVT objective is
\begin{equation}
\mathcal{L}_{\mathrm{IVT}}
=
\mathcal{L}_{\mathrm{text}}
+
\lambda_{\mathrm{pred}}
\mathcal{L}_{\mathrm{pred}},
\label{eq:ivt_loss}
\end{equation}
where we set $\lambda_{\mathrm{pred}}=1$ in all experiments.

\subsection{Design Choices}
\label{subsec:design}
\paragraph{Target representations.}
We compare four future-state target spaces: reconstruction-oriented Flux-VAE latents~\citep{flux2024}, semantic features from DINOv2~\citep{oquab2023dinov2}, and SigLIP2-based features~\citep{tschannen2025siglip} with either an \emph{adaptive} target encoder jointly optimized with the text-generation branch or a \emph{frozen} target encoder. These target spaces differ primarily in two respects. First, reconstruction-oriented latents preserve fine-grained appearance and spatial information required for image reconstruction, whereas semantic representations emphasize higher-level visual concepts and invariances. Second, adaptive targets can evolve together with the model during post-training, while frozen targets provide a stationary prediction space. For all target representations, the target tensor used by $\mathcal{L}_{\mathrm{pred}}$ is detached through $\operatorname{sg}(\cdot)$, such that gradients from the predictive objective do not propagate through the target branch. For frozen encoders, the target-encoder parameters remain fixed throughout training. For the adaptive SigLIP2 variant, the encoder can still be updated through the text-generation objective, allowing the target space to adapt during joint optimization.

\paragraph{Model architectures.}
We study two architectures that differ in how strongly future-state prediction is coupled with language reasoning. In the Dense design, understanding and predictive tokens are processed by the same decoder parameters, so gradients from both $\mathcal{L}_{\mathrm{text}}$ and $\mathcal{L}_{\mathrm{pred}}$ directly update the shared decoder. In the Mixture-of-Experts (MoE) design, understanding and predictive tokens use separate feed-forward experts within each transformer layer. Expert selection is deterministic according to token type rather than controlled by a learned router. Specifically, understanding tokens are processed by the understanding expert, whereas predictive tokens are processed by a separate prediction expert, while the remaining transformer components are shared. This comparison examines whether future-state supervision is most effective when it directly updates the parameters used for language reasoning or when the predictive pathway is partially separated from the understanding pathway.

\paragraph{Prediction horizon.}
We vary the prediction horizon to study the effect of supervising future states at increasing temporal offsets. For a maximum prediction horizon $H$, we define the set of future temporal offsets as $\mathcal{H}_H=\{1,\ldots,H\}$. Frames are sampled at $1~\mathrm{fps}$, so an offset $h\in\mathcal{H}_H$ corresponds to the future frame $I_{t+h}$, occurring $h$ seconds after the final observed frame. Thus, a maximum horizon $H$ supervises the sequence of future states at $+1,\ldots,+H$ seconds relative to the observation boundary.

\paragraph{Data mixture.}
We control the relative amount of language and future-state supervision by varying the mixture of understanding and future-prediction examples packed into each training batch. We compare $1{:}1$, $3{:}1$, and $5{:}1$ understanding-to-prediction mixtures, where larger ratios place greater emphasis on answer-generation supervision.

\paragraph{Training curriculum.}
We compare joint optimization with a two-stage training curriculum. In the two-stage setting, the model is first optimized for future-state prediction and is subsequently fine-tuned using only the downstream understanding objective. This comparison evaluates whether future-state prediction is most effective when learned jointly with language supervision or when introduced as a separate training stage.

\paragraph{Predictive objectives.}
We study two objectives for learning future-state representations. Direct feature regression directly predicts the target representation $Z_h$ from the observed context. For $K=|\mathcal{H}|$ prediction horizons, we optimize
\begin{equation}
\mathcal{L}_{\mathrm{reg}}
=
\frac{1}{KMd}
\sum_{h\in\mathcal{H}}
\left\|
\widehat{Z}_h
-
\operatorname{sg}(Z_h)
\right\|_F^2,
\label{eq:reg_loss}
\end{equation}
where $\|\cdot\|_F$ denotes the Frobenius norm and $\operatorname{sg}(\cdot)$ denotes stop-gradient. The objective therefore corresponds to the mean squared error across prediction horizons, predictive tokens, and feature dimensions. Rectified-flow matching instead learns a conditional velocity field between a noise sample and the target future representation. For each target, we define $Z_h^{(1)}=\operatorname{sg}(Z_h)$, sample $Z_h^{(0)}\sim\mathcal{N}(0,I)$ and $\tau\sim\mathcal{U}(0,1)$, construct $Z_h^{(\tau)}=(1-\tau)Z_h^{(0)}+\tau Z_h^{(1)}$, and define the target velocity as $V_h^\star=Z_h^{(1)}-Z_h^{(0)}$. Conditioned on the observed context, the model predicts $\widehat{V}_h=V_{\theta,\phi}(Z_h^{(\tau)},\tau\mid X_{\leq t},q)$ and minimizes
\begin{equation}
\mathcal{L}_{\mathrm{FM}}
=
\frac{1}{KMd}
\sum_{h\in\mathcal{H}}
\left\|
\widehat{V}_h
-
V_h^\star
\right\|_F^2.
\label{eq:flow_loss}
\end{equation}

\paragraph{Inference.}
At inference time, the model receives only the observed video prefix and textual prompt $(X_{\leq t},q)$ and autoregressively generates the textual response according to $p_{\theta}(y\mid X_{\leq t},q)$. IVT uses the same inference path as the corresponding Answer-Only SFT model and introduces no additional image generation.

%% file: sections/4_experiment.tex
\section{Experiments}
\label{sec:experiments}

\subsection{Experimental Protocol}
\label{sec:setup}

\paragraph{Tasks.}
Given an observed video prefix $X_{\leq t}$ and a textual prompt $q$, the model generates an open-ended textual prediction $y$ about an ongoing or future event. In \textbf{early-event prediction (EEP)}, the observation ends within the annotated action interval $[t_s,t_e]$, such that $t_s<t<t_e$. We evaluate four observation ratios, $\{0.1,0.3,0.5,0.7\}$, measured relative to the duration of the target action~\citep{stergiou2023wisdom}. In \textbf{next-event prediction (NEP)}, the observed prefix ends before the subsequent event begins, satisfying $t\leq t_{\mathrm{next}}-\tau_a$, where the anticipation gap is fixed to $\tau_a=1$ second~\citep{assran2025v,mittal2024can}. We focus on short-horizon anticipation, where immediate motion, object interactions, and local visual dynamics provide informative predictive signals. Long-horizon anticipation depends more strongly on semantic goals, procedural knowledge, and language priors, and is therefore left for future work.

\paragraph{Datasets.}
We train and evaluate on Ego-Exo4D~\citep{grauman2024ego}, Ego4D~\citep{grauman2022ego4d}, and EPIC-KITCHENS-100~\citep{damen2022rescaling}, using their official splits. After preprocessing, the EEP training sets contain $224{,}058$, $83{,}816$, and $268{,}750$ examples, respectively, while the corresponding NEP training sets contain $55{,}943$, $23{,}127$, and $67{,}217$ examples. The EEP validation sets contain $56{,}954$, $45{,}699$, and $38{,}621$ examples, and the NEP validation sets contain $15{,}910$, $11{,}305$, and $9{,}668$ examples. Unless otherwise noted, ablations use a randomly sampled subset of 200 Ego-Exo4D validation examples for early-event prediction. We retain each dataset's native textual supervision: free-form keystep descriptions for Ego-Exo4D, verb--noun action labels for Ego4D, and narrations for EPIC-KITCHENS-100.

\paragraph{Metrics.}
Following prior work~\citep{tan2025kn,liang2022visual,chang2026abductivemllm}, we evaluate open-ended predictions using four metrics: ROUGE-L, METEOR, CIDEr, and BERTScore~\citep{zhangbertscore}. For next-event prediction, we report Top-5 recall to account for the inherent multiplicity of plausible future outcomes, following~\citet{damen2022rescaling,assran2025v,mittal2024can}.

\paragraph{Implementation details.}
We optimize all models with AdamW using $\beta_1=0.9$, $\beta_2=0.95$, $\epsilon=10^{-15}$, and zero weight decay. The learning rate is linearly warmed up for $2{,}000$ steps to $2\times10^{-5}$ and then held constant. Gradients are clipped to a maximum norm of $1.0$, and training uses bfloat16 mixed precision. Frames are sampled at $1~\mathrm{fps}$ and resized to resolutions between $224$ and $448$ pixels. All models are trained for $20{,}000$ optimization steps.
To ensure a controlled comparison with Zebra-CoT~\citep{li2025zebra}, a representative Visual-CoT baseline built on BAGEL-7B~\citep{deng2025bagel}, we initialize the LLM decoder used for visual understanding and future-state prediction from the corresponding pretrained BAGEL checkpoint. BAGEL adopts a Mixture-of-Transformers architecture with separate LLM decoders for visual understanding and generation, totaling approximately $14$B parameters. Our framework instead uses the Dense and MoE architectures described in Section~\ref{subsec:design}.
To isolate the effect of individual design choices, the studies in Sections~\ref{sub:target_repr}, \ref{sub:data_ratio}, \ref{sec:loss_ablation}, and~\ref{sec:curriculum} vary one component at a time from a shared configuration. This configuration uses the MoE decoder with prediction horizon $H=1$, a Flux-VAE target with a flow-matching objective, and a $1{:}1$ understanding-to-prediction mixture. All other factors are held fixed within each study.

\paragraph{Inference efficiency measurement.}
\label{sec:latency}
We use end-to-end wall-clock latency per sample as the primary efficiency metric. Normalizing latency solely by the number of generated text tokens would not provide a controlled comparison because explicit Visual CoT additionally incurs future-frame generation, image decoding, and generated-frame re-encoding. These operations are not reflected in text-decoding throughput.
We profile each method on a single NVIDIA B200 GPU. Measurements use a fixed $200$-example validation subset from each benchmark.
Video frames are loaded before timing; model loading, compilation, and the first warm-up example are excluded. We hold the number and resolution of input frames, prompt format, maximum answer length, and text-decoding configuration fixed across methods. We report both mean and p95 latency, since the latter captures tail behavior that is particularly important for time-sensitive prediction.

\subsection{Motivation: When Does Explicit Visual CoT Help, and at What Cost?}

We adopt Zebra-CoT~\citep{li2025zebra}, a representative Visual CoT baseline, for comparison. Zebra-CoT first generates a plausible future frame as an explicit visual thought and then conditions the final textual prediction on the generated frame. Table~\ref{tab:vcot-prelim} compares this pipeline with Answer-Only SFT under a controlled setting using the same backbone and fine-tuning data. To disentangle the value of future visual context from errors introduced by visual generation, we additionally evaluate an \emph{oracle-frame} variant: at the same model checkpoint, the generated frame is replaced with the corresponding ground-truth future frame. Although this oracle is not available in practice, it provides a diagnostic upper bound on how effectively the model can exploit accurate future visual evidence.

\begin{table}[t]
\centering
\scriptsize
\setlength{\tabcolsep}{2pt}
\renewcommand{\arraystretch}{1.25}
\newcommand{\gimp}[1]{{\scriptsize\textcolor{gain}{($+$#1\%)}}}
\newcommand{\dimp}[1]{{\scriptsize\textcolor{loss}{($-$#1\%)}}}
\newcommand{\slow}[1]{{\scriptsize\textcolor{loss}{(#1$\times$)}}}
\caption{\textbf{Accuracy--efficiency trade-off of Visual CoT.} For each task, we report performance averaged across the validation subsets of Ego-Exo4D, Ego4D, and EPIC-KITCHENS-100. Conditioning on a generated future frame improves early-event prediction but provides limited benefit for next-event prediction, while substantially increasing mean inference latency. Replacing the generated frame with the ground-truth future frame improves both tasks, revealing a considerably higher upper bound when accurate future visual context is available. This gap suggests that the effectiveness of Visual CoT depends strongly on the fidelity of the generated future state, as intermediate images that fail to preserve plausible scene dynamics or task-relevant visual information may provide weak or misleading conditioning signals.}
\label{tab:vcot-prelim}
\begin{tabular}{@{}ll cc cccc@{}}
\toprule
& & \multicolumn{2}{c}{Latency ($\downarrow$)} & \multicolumn{4}{c}{Answer Quality ($\uparrow$)} \\
\cmidrule(lr){3-4}\cmidrule(l){5-8}
Task & Method & Mean & P95 & ROUGE-L & BERTScore & METEOR & CIDEr \\
\midrule
\multirow{3}{*}{Early Event Pred.}
& Answer-Only SFT & 1.07 & 1.79 & 34.4 & 41.3 & 24.3 & 1.43 \\
& Visual CoT (generated frame) & 6.56\,\slow{6.2} & 8.72\,\slow{4.9} & 36.4\,\gimp{5.9} & 43.2\,\gimp{4.5} & 29.8\,\gimp{22.4} & 1.69\,\gimp{17.9} \\
& Visual CoT (oracle frame) & --- & --- & 41.9\,\gimp{21.6} & 48.3\,\gimp{17.0} & 34.2\,\gimp{40.6} & 2.05\,\gimp{43.4} \\
\midrule
\multirow{3}{*}{Next Event Pred.}
& Answer-Only SFT & 1.32 & 2.05 & 47.2 & 50.2 & 35.3 & 2.14 \\
& Visual CoT (generated frame) & 6.55\,\slow{5.0} & 7.12\,\slow{3.5} & 46.8\,\dimp{0.8} & 49.9\,\dimp{0.7} & 36.9\,\gimp{4.3} & 2.13\,\dimp{0.6} \\
& Visual CoT (oracle frame) & --- & --- & 50.4\,\gimp{6.8} & 52.5\,\gimp{4.6} & 40.4\,\gimp{14.2} & 2.38\,\gimp{11.3} \\
\bottomrule
\end{tabular}
\end{table}

\begin{observation}
For early-event prediction, Visual CoT improves all four metrics, including ROUGE-L by $5.9\%$ and CIDEr by $17.9\%$. For next-event prediction, however, its effect is negligible or negative on three of the four metrics, with only METEOR improving by $4.3\%$. These modest and inconsistent gains come with substantial computational overhead. Specifically, mean latency increases by $6.2\times$ for early-event prediction and $5.0\times$ for next-event prediction.
In contrast, conditioning the same model on the ground-truth future frame improves every metric on both tasks, yielding ROUGE-L gains of $21.6\%$ for early-event prediction and $6.8\%$ for next-event prediction.
\end{observation}

\begin{takeaway}
\textbf{The primary bottleneck of explicit Visual CoT is not the value of future visual context, but the cost and fidelity of externalizing that context as pixels.} Accurate future frames substantially improve both early- and next-event prediction, confirming that future visual state contains useful evidence for proactive reasoning. However, self-generated frames recover only a fraction of this benefit for early-event prediction and provide no consistent gain for next-event prediction, where the target event lies entirely beyond the observed video prefix. At the same time, generating and reprocessing the intermediate frame increases mean inference latency by approximately $5$--$6\times$.
Those experiments identify a promising alternative by using future-state supervision as a training signal while removing pixel-space visual generation from the inference path.
\end{takeaway}

\subsection{Which Future Representation Should the Model Predict?}
\label{sub:target_repr}

\begin{figure}[t]
\centering
\includegraphics[width=\linewidth]{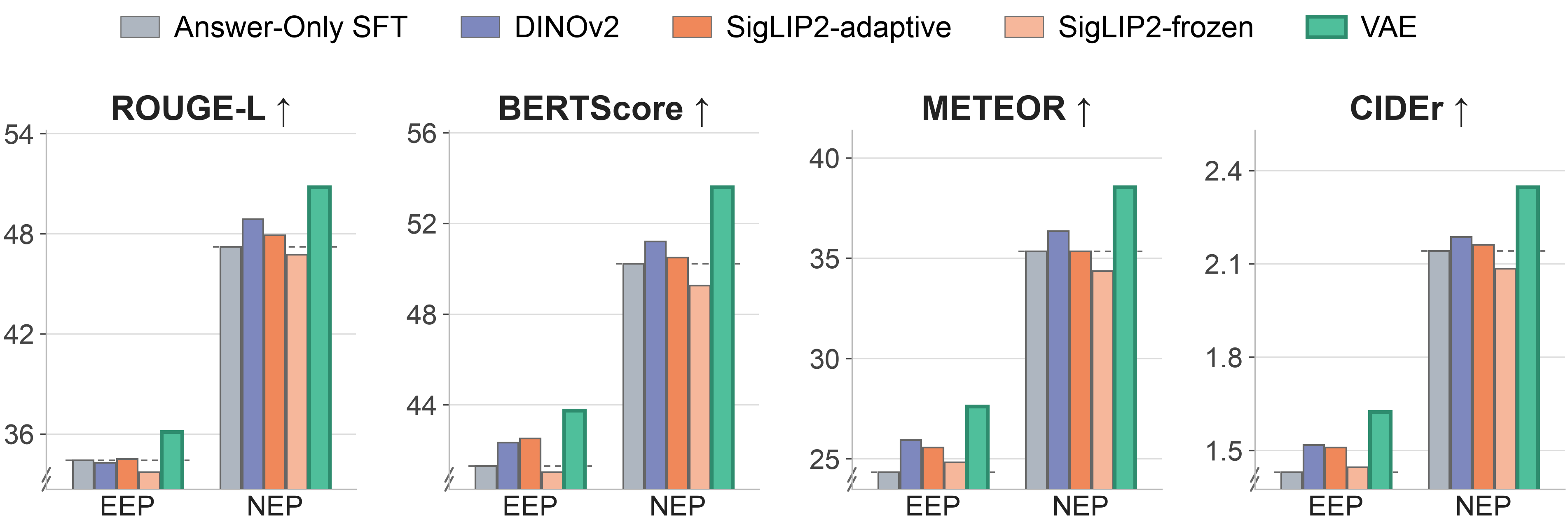}
\caption{\textbf{Comparison of different target representations.}
We report ROUGE-L, BERTScore, METEOR, and CIDEr, averaged across the three benchmarks for early-event and next-event prediction. The dashed line denotes the Answer-Only SFT baseline. Predicting a Flux-VAE latent consistently achieves the strongest performance across all metrics and both tasks, whereas semantic targets provide smaller and less consistent gains.}
\label{fig:targets}
\end{figure}

We first investigate \emph{what} the model should predict as its future-state supervision.
Figure~\ref{fig:targets} compares four target representations: reconstruction-oriented latents from Flux-VAE; semantic features from DINOv2; and features from a SigLIP2-style ViT encoder that is either jointly optimized with the text-generation branch (\texttt{SigLIP2-adaptive}) or kept frozen (\texttt{SigLIP2-frozen}). We compare all variants with Answer-Only SFT, which receives no future-embedding supervision. Results are averaged across the three benchmarks for both early-event and next-event prediction.

\begin{observation}
The choice of target representation has a substantial effect on whether future-state prediction benefits downstream reasoning. Flux-VAE brings substantial performance improvements on both tasks. In contrast, the semantic targets yield smaller and less consistent gains. DINOv2 improves three of the four early-event metrics and all four next-event metrics, but remains consistently below Flux-VAE. The adaptive SigLIP2 target generally outperforms its frozen counterpart.

\end{observation}

\begin{takeaway}
\textbf{The effectiveness of future-state supervision depends on the choice of target representation.}
Both Flux-VAE and DINOv2 targets improve over Answer-Only SFT, while Flux-VAE delivers the strongest and most consistent gains across tasks and metrics. This suggests that the benefit of future-state prediction depends not simply on adding an auxiliary visual objective, but on whether the target representation exposes structure that the model can use to anticipate how the scene will evolve.
DINOv2 features remain beneficial, indicating that high-level semantic structure alone can still provide useful predictive supervision.
\end{takeaway}

\subsection{How Much Predictive Supervision Is Needed?}
\label{sub:data_ratio}

\begin{figure}[t]
\centering
\includegraphics[width=\linewidth]{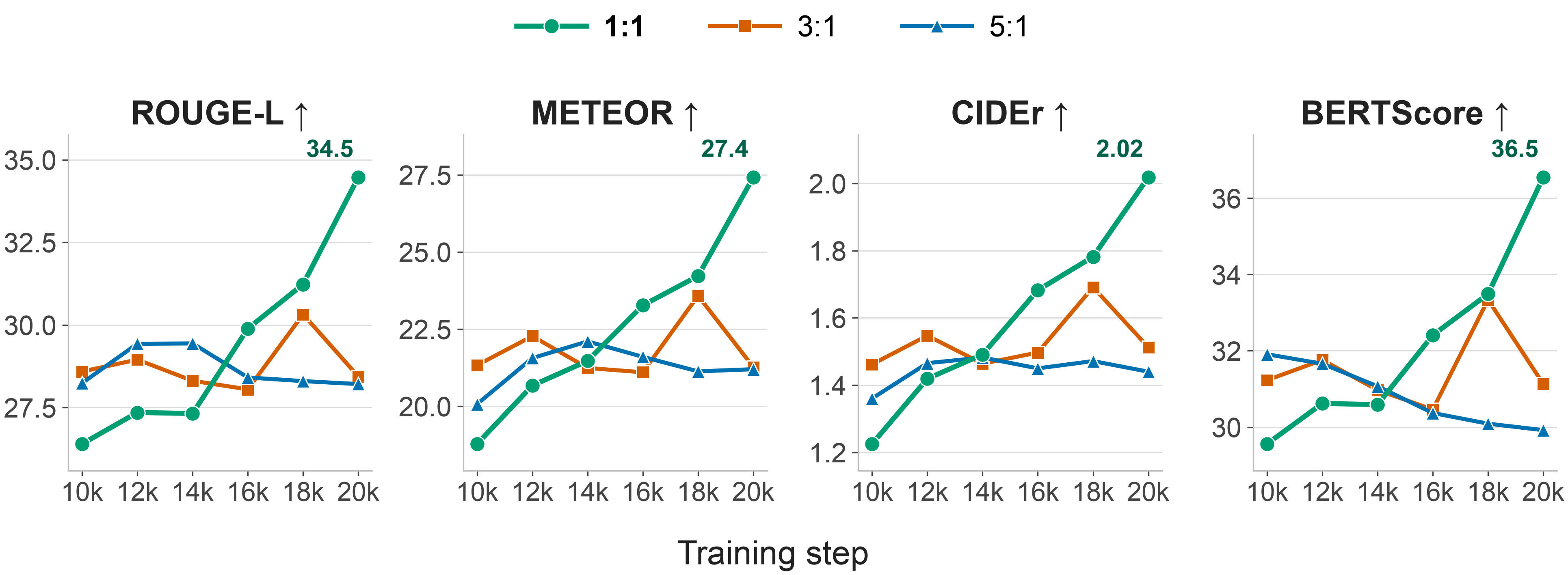}
\caption{\textbf{Effect of the understanding-to-prediction data-mixture ratio} on Ego-Exo4D early-event prediction.
We report performance throughout training for three sampling ratios. Understanding-heavy mixtures ($3{:}1$ and $5{:}1$) improve more rapidly at early checkpoints but subsequently plateau or decline. In contrast, the balanced $1{:}1$ mixture improves steadily and achieves the strongest result on all four metrics at 20k steps.}
\label{fig:dratio}
\end{figure}

We next examine how frequently future-prediction examples should be interleaved with understanding examples during joint training. Figure~\ref{fig:dratio} compares three understanding-to-prediction sampling ratios: $1{:}1$, $3{:}1$, and $5{:}1$, where larger ratios allocate a greater fraction of updates to the understanding objective. This ablation tests whether next-embedding prediction is effective as occasional auxiliary supervision or instead requires sustained exposure throughout training.

\begin{observation}
The data-mixture ratio changes not only final performance but also the optimization trajectory. The understanding-heavy $3{:}1$ and $5{:}1$ mixtures perform better at the earliest checkpoints, suggesting that allocating more updates to answer supervision accelerates initial task adaptation. However, both configurations converge after approximately 12--18k steps and subsequently fluctuate or decline. Their final performance remains close to, or below, their early peaks.
The balanced $1{:}1$ mixture exhibits the opposite pattern. Although it begins below the understanding-heavy variants, it continues to improve across later checkpoints and achieves the best result on every metric at 20k steps, including ROUGE-L of $34.5$, METEOR of $27.4$, CIDEr of $2.02$, and BERTScore of $36.5$.
\end{observation}

\begin{takeaway}
\textbf{Future-state prediction should be treated as a primary training objective rather than a sparsely applied auxiliary regularizer.}
Oversampling understanding examples yields faster early progress, but this advantage is transient. Reducing the frequency of predictive updates causes performance to saturate well below that of the balanced mixture. The $1{:}1$ result suggests that sustained optimization of both objectives is necessary for the predictive signal to influence the representations used for proactive reasoning, rather than being overwritten by task-specific answer supervision. We therefore adopt a balanced $1{:}1$ understanding-to-prediction ratio in the remaining experiments.
\end{takeaway}

\subsection{How Should Future Representations Be Predicted?}
\label{sec:loss_ablation}

\begin{figure}[t]
\centering
\includegraphics[width=\linewidth]{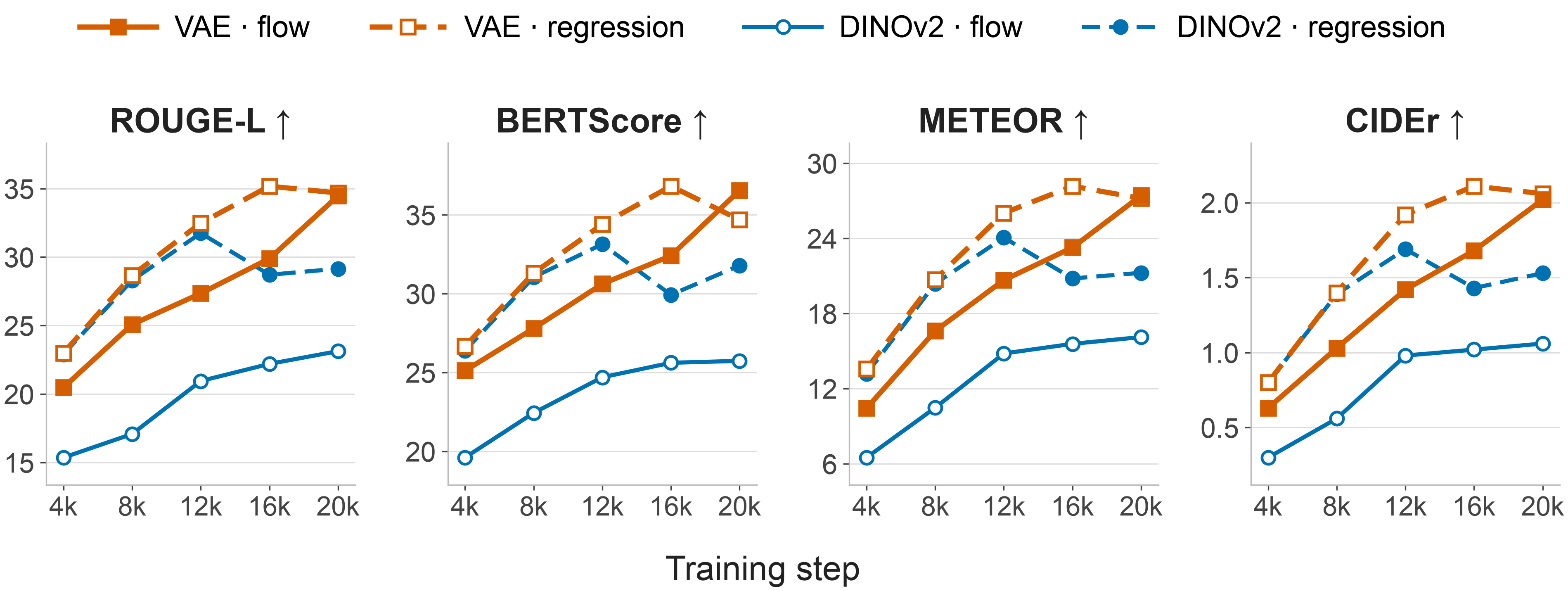}
\caption{\textbf{Comparison of predictive objectives} on Ego-Exo4D early-event prediction.
We compare rectified-flow matching and direct feature regression for VAE and DINOv2 future-state targets. Direct regression substantially outperforms flow matching for DINOv2 and is also competitive with flow matching for VAE across most checkpoints and metrics.}
\label{fig:lossablation}
\end{figure}

Having identified the target representation and data-mixture ratio, we next examine \emph{how} future-state prediction should be supervised. We compare two predictive objectives. The first is \emph{rectified-flow matching}, in which the prediction head estimates the velocity field from a noisy intermediate state toward the target representation at a sampled flow timestep. The second is \emph{direct feature regression}, in which the model deterministically predicts the clean target representation. Figure~\ref{fig:lossablation} compares these objectives for the VAE and DINOv2 targets.

\begin{observation}
The preferred predictive objective depends strongly on the target representation. For DINOv2, direct regression provides a consistent advantage over flow matching across all four metrics.
For the VAE target, the difference is considerably smaller. Direct regression leads throughout training on ROUGE-L and CIDEr and at most checkpoints on BERTScore and METEOR. Flow matching becomes competitive at the final checkpoint, slightly surpassing regression on BERTScore ($36.5$ vs.\ $34.7$) and METEOR ($27.4$ vs.\ $27.2$).
These results reveal an interaction between the representation and the predictive objective. Direct regression is decisively preferable for DINOv2, whereas VAE latents can be learned effectively with either objective.
\end{observation}

\begin{takeaway}
\textbf{The target representation and predictive objective should be designed jointly rather than optimized in isolation.}
Direct regression substantially outperforms flow matching for DINOv2 and achieves similar performance to flow matching for VAE. Notably, although BAGEL performs visual generation in VAE latent space using rectified flow, direct regression achieves similar downstream reasoning performance. This suggests that the most suitable predictive objective is determined less by the formulation originally used to train the visual-generation branch and more by the properties of the target representation itself, such as its scale, normalization, dimensionality, and predictability from the observed context.
\end{takeaway}

\subsection{Which Training Curriculum Best Integrates Predictive Supervision?}
\label{sec:curriculum}

We next examine whether future-state prediction should be introduced as a separate training stage or optimized jointly with the downstream task. We compare three curricula on Ego-Exo4D early-event prediction: (i) \textbf{Answer-Only SFT}, which optimizes only the textual answer objective; (ii) \textbf{Two-Stage Training}, which first performs next-embedding prediction for 10k steps and then fine-tunes exclusively on the understanding task; and (iii) \textbf{Joint Training}, which maintains both the next-embedding-prediction and answer-generation objectives throughout post-training.

\input{sections/tab_two_stage}

\begin{observation}
Joint training achieves the strongest performance on every metric, improving over Answer-Only SFT by $12.6\%$ in ROUGE-L and $26.3\%$ in CIDEr. In contrast, two-stage training underperforms Answer-Only SFT on \emph{every} metric by $6.5\%$ in ROUGE-L, $5.6\%$ in BERTScore, $5.5\%$ in METEOR and $3.5\%$ in CIDEr. Therefore, predictive training followed by task-only fine-tuning is worse than no predictive training.

\end{observation}

\begin{takeaway}
\textbf{Next visual state prediction is most effective when it is jointly trained with text generation throughout post-training.}
Joint optimization allows predictive representations to remain directly aligned with the downstream language objective, whereas separating the objectives into sequential stages substantially weakens their transfer.
Thus, the benefit of next-embedding prediction should not be interpreted as a generic predictive training stage that can simply precede task adaptation. It arises from continuously coupling visual-dynamics supervision with the representations used to generate the answer.

\end{takeaway}

\subsection{How Should Future Prediction Be Structured?}
\label{sec:arch}

\begin{figure}[t]
\centering
\includegraphics[width=\linewidth]{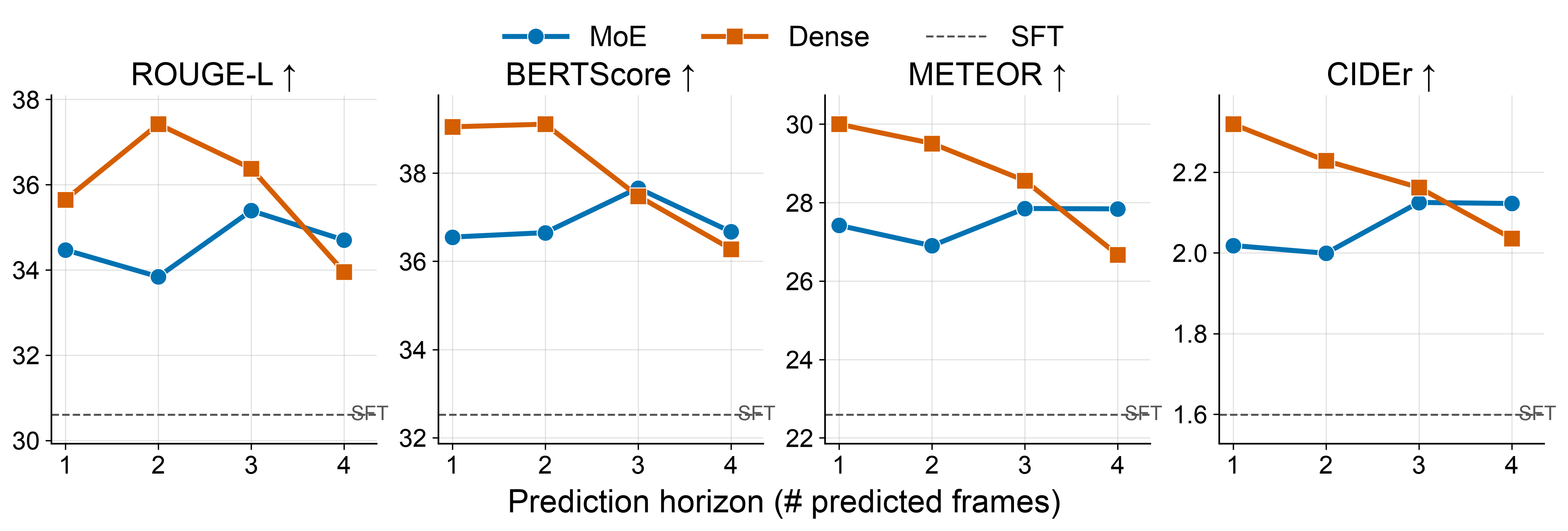}
\caption{\textbf{Interaction between decoder architecture and prediction horizon} on Ego-Exo4D early-event prediction.
We report the experimental results for each prediction horizon \(H\), defined as the number of predicted future frames. The dashed line denotes the Answer-Only SFT baseline. Both architectures outperform it across all horizons. The shared Dense decoder performs best at short horizons, whereas the MoE architecture is comparatively less sensitive to horizon and reaches its strongest results around \(H=3\).}
\label{fig:arch-horizon}
\end{figure}

We finally examine whether future-state prediction and language understanding should be optimized through shared or task-specific decoder parameters. Figure~\ref{fig:arch-horizon} compares two architectures on Ego-Exo4D early-event prediction: a \textbf{Dense} model, in which both objectives update a single shared decoder, and a \textbf{Mixture-of-Experts (MoE)} model, in which predictive and language objectives are assigned to separate experts. We evaluate both architectures across different prediction horizons to determine how parameter sharing interacts with the amount of future-state supervision.

\begin{observation}
Both architectures outperform Answer-Only SFT across all four metrics and every evaluated horizon.
However, the magnitude of this benefit depends strongly on the architecture. At short horizons (\(H=1,2\)), the shared Dense model consistently outperforms MoE on all four metrics. For example, at \(H=2\), Dense achieves \(37.4\) ROUGE-L, compared with \(33.8\) for MoE.
The two architectures also exhibit different sensitivity to prediction horizon. Dense performs best with one or two predicted frames and subsequently declines as \(H\) increases. MoE is comparatively flatter: its performance generally peaks around \(H=3\) and changes only modestly between \(H=3\) and \(H=4\). This interaction suggests that dense parameter sharing is particularly effective when the predictive target is sufficiently constrained, but becomes less advantageous as the prediction task grows more uncertain.
\end{observation}

\begin{takeaway}
\textbf{The dense architecture most effectively transfers predictive supervision to language reasoning when the future target is short and well constrained.}
A shared decoder allows gradients from next-embedding prediction to directly shape the representations used for answer generation, which is consistent with its clear advantage at \(H=1\) and \(H=2\). As the horizon increases, however, future prediction becomes more ambiguous, and tightly coupling this objective to the language decoder may introduce less task-aligned supervision. Separating the objectives through MoE appears to reduce this sensitivity, although it does not match the strongest short-horizon Dense configuration.

\end{takeaway}

\subsection{Main Results: Quality and Efficiency}
\label{sec:main_results}

\input{sections/tab_sota}

Tables~\ref{tab:sota_egoexo}--\ref{tab:sota_ek100} compare IVT with four representative open-source video MLLMs: LLaVA-NeXT-Video-7B~\citep{zhang2024llavanext-video}, VideoLLaMA3-7B~\citep{zhang2025videollama}, Qwen2.5-VL-7B~\citep{bai2025qwen25vltechnicalreport}, and Qwen3-VL-8B~\citep{bai2025qwen3}. The Qwen models provide strong general-purpose vision-language foundation-model baselines, while LLaVA-NeXT-Video and VideoLLaMA3 represent video-specialized MLLMs. All models are fine-tuned on the corresponding target dataset. Because the external models differ in backbone, pretraining data, and optimization recipe, we treat them as reference points rather than as strictly controlled comparisons.
Within the BAGEL-7B-based block, \emph{Answer-Only SFT} denotes the matched backbone trained only with the textual prediction objective, without future-state supervision. Text CoT, Visual CoT, and IVT share the same initialization for the understanding branch. Accordingly, \textbf{bold} and \underline{underlined} values denote the best and second-best results only within this controlled BAGEL-7B-based comparison.

For the Text CoT baseline, we use Qwen3-VL-30B-A3B-Thinking~\citep{bai2025qwen3} as a teacher to generate intermediate reasoning traces conditioned on the input frames and ground-truth event descriptions. These traces are then used as supervision for post-training the BAGEL-7B-based Text CoT model.
For IVT, we use the configurations selected from the architecture--horizon study in Figure~\ref{fig:arch-horizon}, with $H=1$ for Dense and $H=3$ for MoE. We use Flux-VAE as the target representation, optimize it with the flow-matching objective, and set the understanding-to-prediction data-mixture ratio to $1{:}1$. These configurations are fixed across all six dataset--task settings.

\begin{observation}
Relative to the Answer-Only SFT baseline, Visual CoT improves four of the six dataset--task settings, while Text CoT generally degrades performance. In contrast, IVT improves all four metrics across all six settings, showing consistent gains for both partially observed actions and future events that have not yet begun.
Compared with Visual CoT, IVT performs better in four of the six settings and wins on all three next-event prediction benchmarks. Visual CoT remains stronger on two early-event prediction benchmarks, exceeding IVT by $1.3$ ROUGE-L points on Ego4D and $1.2$ points on EPIC-KITCHENS-100. Compared with external baselines, IVT is competitive with state-of-the-art models at the 7B scale, while still trailing the larger Qwen3-VL-8B model.
\end{observation}

\begin{takeaway}
\textbf{IVT improves the quality--efficiency frontier by moving future-state prediction from inference-time computation to training-time supervision.}
Unlike explicit Visual CoT, IVT retains the Answer-Only SFT inference pathway while achieving stronger overall performance. The advantage is particularly clear for next-event prediction, where IVT improves the average ROUGE-L from $45.9$ to $49.7$ over Visual CoT across the three benchmarks. These results suggest that explicit future-frame generation is not necessary to obtain the benefits of visual foresight. Instead, predictive world modeling can be internalized during training to support more accurate and efficient proactive reasoning.
\end{takeaway}

%% file: sections/tab_two_stage.tex
\begin{table}[t]
\centering
\small
\setlength{\tabcolsep}{5pt}
\newcommand{\gimp}[1]{{\scriptsize\textcolor{gain}{($+$#1\%)}}}
\newcommand{\dimp}[1]{{\scriptsize\textcolor{loss}{($-$#1\%)}}}
\caption{\textbf{Joint optimization is more effective than sequential training.}
We report the results on Ego-Exo4D early-event prediction. Relative changes are computed against the Answer-Only SFT baseline. Two-stage training underperforms Answer-Only SFT on every metric, whereas joint training achieves the strongest result across all metrics. Best value in each column is shown in \textbf{bold}.}
\label{tab:curriculum}
\begin{tabular}{@{}lllll@{}}
\toprule
Method & ROUGE-L & BERTScore & METEOR & CIDEr \\
\midrule
Answer-Only SFT & 30.6 & 32.5 & 22.6 & 1.60 \\
Two-Stage Training & 28.6\,\dimp{6.5} & 30.7\,\dimp{5.6} & 21.3\,\dimp{5.5} & 1.54\,\dimp{3.5} \\
\textbf{Joint Training} & \textbf{34.5}\,\gimp{12.6} & \textbf{36.5}\,\gimp{12.4} & \textbf{27.4}\,\gimp{21.4} & \textbf{2.02}\,\gimp{26.3} \\
\bottomrule
\end{tabular}
\end{table}

%% file: sections/tab_sota.tex
\begin{table}[!t]
\centering
\small
\setlength{\tabcolsep}{3pt}
\caption{\textbf{Results on the full Ego-Exo4D validation set.}
We evaluate early-event prediction with top-1 decoding and next-event prediction with top-5 decoding using ROUGE-L, BERTScore, METEOR, and CIDEr. \emph{Answer-Only SFT} uses the same matched backbone and training setup as IVT but removes future-state prediction, retaining only the textual prediction objective. External models are included as reference points, while \textbf{bold} and \underline{underlined} values denote the best and second-best results within the controlled BAGEL-7B-based block.}
\label{tab:sota_egoexo}
\begin{tabular}{@{}l cccc cccc@{}}
\toprule
 & \multicolumn{4}{c}{\textbf{Early Event Pred.} (top-1)} & \multicolumn{4}{c}{\textbf{Next Event Pred.} (top-5)} \\
\cmidrule(lr){2-5}\cmidrule(lr){6-9}
Method & ROUGE-L & BERTScore & METEOR & CIDEr & ROUGE-L & BERTScore & METEOR & CIDEr \\
\midrule
\multicolumn{9}{@{}l}{\textit{External models fine-tuned on the same data}} \\
LLaVA-NeXT-Video-7B & 24.3 & 5.5 & 17.4 & 1.11 & 44.2 & 42.3 & 33.7 & 2.40 \\
VideoLLaMA3-7B & 14.6 & 17.4 & 7.7 & 0.31 & 30.2 & 26.3 & 19.6 & 1.30 \\
Qwen2.5-VL-7B & 34.9 & 36.0 & 26.6 & 1.75 & 48.0 & 45.9 & 37.9 & 2.77 \\
Qwen3-VL-8B & 37.0 & 38.1 & 28.5 & 1.90 & 52.1 & 49.5 & 42.0 & 3.09 \\
\midrule
\multicolumn{9}{@{}l}{\textit{BAGEL-7B-based methods}} \\
Answer-Only SFT & 28.8 & 29.8 & 21.3 & 1.40 & 46.3 & 45.0 & 35.6 & 2.59 \\
Text-CoT & 24.6 & 26.1 & 17.1 & 1.10 & 32.6 & 33.0 & 22.0 & 1.48 \\
Visual-CoT & 32.7 & 33.2 & 25.4 & 1.69 & 42.6 & 41.1 & 31.6 & 2.22 \\
IVT (Ours, MoE) & \textbf{33.3} & \textbf{34.2} & \textbf{25.7} & \textbf{1.74} & \textbf{48.5} & \underline{46.6} & \textbf{37.9} & \underline{2.75} \\
\textbf{IVT (Ours, Dense)} & \underline{33.2} & \underline{34.0} & \underline{25.5} & \underline{1.72} & \underline{48.2} & \textbf{46.8} & \underline{37.8} & \textbf{2.76} \\
\bottomrule
\end{tabular}

\vspace{1.6em}

\caption{\textbf{Results on the full Ego4D validation set.}
We evaluate early-event prediction with top-1 decoding and next-event prediction with top-5 decoding using ROUGE-L, BERTScore, METEOR, and CIDEr. \emph{Answer-Only SFT} uses the same matched backbone and training setup as IVT but removes future-state prediction, retaining only the textual prediction objective. External models are included as reference points, while \textbf{bold} and \underline{underlined} values denote the best and second-best results within the controlled BAGEL-7B-based block.}
\label{tab:sota_ego4d}
\begin{tabular}{@{}l cccc cccc@{}}
\toprule
 & \multicolumn{4}{c}{\textbf{Early Event Pred.} (top-1)} & \multicolumn{4}{c}{\textbf{Next Event Pred.} (top-5)} \\
\cmidrule(lr){2-5}\cmidrule(lr){6-9}
Method & ROUGE-L & BERTScore & METEOR & CIDEr & ROUGE-L & BERTScore & METEOR & CIDEr \\
\midrule
\multicolumn{9}{@{}l}{\textit{External models fine-tuned on the same data}} \\
LLaVA-NeXT-Video-7B & 32.9 & 33.4 & 22.3 & 0.98 & 52.6 & 63.9 & 39.5 & 1.91 \\
VideoLLaMA3-7B & 25.5 & 41.5 & 15.3 & 0.57 & 41.7 & 53.1 & 28.5 & 1.27 \\
Qwen2.5-VL-7B & 33.1 & 49.5 & 22.3 & 0.97 & 51.1 & 62.5 & 37.8 & 1.80 \\
Qwen3-VL-8B & 34.3 & 50.5 & 23.7 & 1.06 & 52.5 & 62.1 & 38.9 & 1.83 \\
\midrule
\multicolumn{9}{@{}l}{\textit{BAGEL-7B-based methods}} \\
Answer-Only SFT & 33.3 & 49.2 & 22.6 & 0.99 & 51.9 & 63.7 & 39.0 & 1.87 \\
Text-CoT & 31.7 & 50.6 & 21.3 & 0.98 & 40.9 & 55.6 & 27.8 & 1.29 \\
Visual-CoT & \textbf{36.1} & \textbf{52.6} & \textbf{25.1} & \textbf{1.13} & 52.2 & 63.0 & 39.3 & \underline{1.88} \\
IVT (Ours, MoE) & 34.0 & 50.4 & 23.2 & 1.03 & \underline{52.5} & \underline{64.4} & \underline{39.5} & \textbf{1.90} \\
\textbf{IVT (Ours, Dense)} & \underline{34.8} & \underline{50.8} & \underline{23.8} & \underline{1.07} & \textbf{52.7} & \textbf{64.5} & \textbf{39.7} & \textbf{1.90} \\
\bottomrule
\end{tabular}
\end{table}

\begin{table}[!t]
\centering
\small
\setlength{\tabcolsep}{3pt}
\caption{\textbf{Results on the full EPIC-KITCHENS-100 validation set.}
We evaluate early-event prediction with top-1 decoding and next-event prediction with top-5 decoding using ROUGE-L, BERTScore, METEOR, and CIDEr. \emph{Answer-Only SFT} uses the same matched backbone and training setup as IVT but removes future-state prediction, retaining only the textual prediction objective. External models are included as reference points, while \textbf{bold} and \underline{underlined} values denote the best and second-best results within the controlled BAGEL-7B-based block.}
\label{tab:sota_ek100}
\begin{tabular}{@{}l cccc cccc@{}}
\toprule
 & \multicolumn{4}{c}{\textbf{Early Event Pred.} (top-1)} & \multicolumn{4}{c}{\textbf{Next Event Pred.} (top-5)} \\
\cmidrule(lr){2-5}\cmidrule(lr){6-9}
Method & ROUGE-L & BERTScore & METEOR & CIDEr & ROUGE-L & BERTScore & METEOR & CIDEr \\
\midrule
\multicolumn{9}{@{}l}{\textit{External models fine-tuned on the same data}} \\
LLaVA-NeXT-Video-7B & 36.8 & 30.3 & 27.2 & 1.49 & 46.3 & 46.3 & 33.1 & 1.98 \\
VideoLLaMA3-7B & 15.6 & 19.9 & 9.8 & 0.44 & 24.3 & 22.2 & 13.3 & 0.63 \\
Qwen2.5-VL-7B & 35.8 & 38.6 & 26.0 & 1.39 & 49.2 & 47.9 & 35.6 & 2.12 \\
Qwen3-VL-8B & 39.5 & 41.9 & 29.1 & 1.59 & 52.9 & 51.8 & 39.7 & 2.37 \\
\midrule
\multicolumn{9}{@{}l}{\textit{BAGEL-7B-based methods}} \\
Answer-Only SFT & 36.5 & 40.0 & 26.7 & 1.45 & 45.5 & 46.4 & 32.5 & 1.90 \\
Text-CoT & 26.8 & 31.2 & 19.0 & 0.92 & 32.3 & 33.7 & 20.6 & 1.12 \\
Visual-CoT & \textbf{40.2} & \textbf{42.4} & \textbf{29.8} & \textbf{1.67} & 42.8 & 44.0 & 30.0 & 1.75 \\
IVT (Ours, MoE) & \underline{39.0} & \underline{42.3} & \underline{29.0} & \underline{1.59} & \textbf{49.1} & \textbf{48.9} & \textbf{35.5} & \textbf{2.08} \\
\textbf{IVT (Ours, Dense)} & \underline{39.0} & 41.9 & 28.9 & \underline{1.59} & \underline{48.3} & \underline{48.4} & \underline{34.5} & \underline{2.01} \\
\bottomrule
\end{tabular}
\end{table}

%% file: sections/2_related.tex
\section{Related Work}

\paragraph{Proactive Video Reasoning.}
Proactive video reasoning seeks to infer ongoing or future events, underlying goals, and plausible outcomes from incomplete visual observations. Following the definition in~\citet{zhao2021review}, proactive video reasoning comprises two primary tasks. \emph{Early-event prediction} requires recognizing an ongoing action before it is completed~\citep{sadegh2017encouraging,wang2019progressive,stergiou2023wisdom}, whereas \emph{next-event prediction} requires anticipating an action that has not yet begun and may occur several seconds into the future~\citep{furnari2020rolling,girdhar2021anticipative,gong2022future}. More recent approaches incorporate the semantic and commonsense priors of large language models to generate plausible long-horizon futures~\citep{mittal2024can,zhao2024antgpt}. Beyond predicting \emph{what} will happen, ProactiveVideoQA~\citep{wang2025proactivevideoqa} evaluates whether video MLLMs can determine \emph{when} to respond during streaming video, extending evaluation from response correctness to temporally appropriate interaction.
\citet{wang2025fostering} formulate next-event prediction as a self-supervised video-to-text objective in which an MLLM observes the first part of a video and predicts a textual description of its unseen continuation. Their analysis shows that supervised fine-tuning provides a strong and efficient baseline, whereas reinforcement learning substantially improves performance on the targeted future-reasoning benchmark but can reduce general video-understanding performance. Video-as-Answer~\citep{cheng2026video} further extends next-event prediction from textual descriptions to generated future videos, directly visualizing the predicted continuation. These methods demonstrate the value of predictive supervision but express the anticipated future through language or computationally expensive visual outputs. In contrast, our method learns future dynamics internally by jointly predicting the next latent visual representation and the future textual state.

\paragraph{Visual Chain-of-Thought.}
Chain-of-thought reasoning~\citep{wei2022chain} has been extended beyond language by introducing explicit visual intermediates into the reasoning trajectory. Visual CoT~\citep{shao2024visual} identifies and revisits question-relevant regions through intermediate bounding boxes; Visual Sketchpad~\citep{hu2024visual} enables models to draw auxiliary visual marks and invoke specialist vision tools; and Visualization-of-Thought methods externalize intermediate spatial states as textual or generated visualizations~\citep{wu2024mind,li2025imagine}. For video reasoning, Video-of-Thought~\citep{fei2024video} structures reasoning from fine-grained spatiotemporal perception to high-level cognition using scene-graph representations. Zebra-CoT~\citep{li2025zebra} provides large-scale interleaved image--text reasoning trajectories for training unified multimodal models, while CoT-VLA~\citep{zhao2025cot} predicts future images as visual subgoals before decoding robotic actions.
Although explicit visual reasoning can improve spatial, temporal, and embodied reasoning, it introduces substantial inference overhead through image-token decoding, repeated visual-model calls, or both. Our approach addresses this limitation in the temporal domain. Rather than generating images, sketches, or other explicit visual thoughts, IVT internalizes predictive visual reasoning within the model's latent computation, retaining the benefits of visual imagination while substantially reducing the inference cost.

\paragraph{Latent Visual Reasoning and Next-Embedding Prediction.}
Joint-embedding predictive architectures demonstrate that predicting semantic representations in latent space can provide a more abstract learning signal than reconstructing raw pixels~\citep{assran2023self,bardes2024revisiting,assran2025v,xu2025next,li2025latent,chen2025vl,zhu2023stmt,wang2025fostering,tong2026beyond}. I-JEPA~\citep{assran2023self} predicts masked image representations from visible context, while V-JEPA~\citep{bardes2024revisiting} extends this principle to video by predicting masked spatiotemporal representations. V-JEPA~2~\citep{assran2025v} further scales latent video prediction to large-scale video understanding, action anticipation, and action-conditioned planning, demonstrating that predictive representations can encode physical dynamics without explicitly synthesizing future pixels.
Related ideas have recently been incorporated into vision and vision-language models. NEPA~\citep{xu2025next} trains causal vision transformers solely through next-embedding prediction. Collectively, these studies show that embedding-space prediction can support visual representation learning, multimodal reasoning, and efficient vision-language inference.
Our approach differs from these methods by jointly predicting the next latent visual representation and the next textual event state during MLLM post-training, explicitly coupling visual dynamics with semantic anticipation.

\paragraph{Unified Understanding and Generation Models.}
Unified multimodal models seek to support visual understanding and generation within a shared architecture~\citep{sun2024emu,team2024chameleon,xie2025show,zhu2024open,zhou2025transfusion,wu2025janus,tian2026unigen,tian2025unigen}. Emu~\citep{sun2024emu} and Chameleon~\citep{team2024chameleon} autoregressively model interleaved image and text sequences, while Show-o~\citep{xie2025show} combines autoregressive text modeling with discrete visual generation in a single transformer. Transfusion~\citep{zhou2025transfusion} instead combines next-token prediction for language with diffusion-based prediction for continuous visual representations. Janus~\citep{wu2025janus} further shows that separating the visual encoding pathways for understanding and generation alleviates representational conflicts while retaining a shared multimodal transformer.
These models primarily employ their generative pathways to produce observable visual outputs. Our objective is fundamentally different: we use the visual generation pathway as an internal predictive mechanism. Specifically, IVT forecasts future visual representations from partial video evidence and uses predictive supervision to improve anticipatory language reasoning, without decoding them into images at inference time.

%% file: sections/5_conclusion.tex
\section{Conclusion}

We introduced \emph{Internalized Visual Thinking} (IVT), a post-training framework for proactive video reasoning that learns to predict future visual states during training without explicitly generating them at inference time. By jointly supervising future latent representations and textual predictions, IVT transfers visual foresight from an expensive intermediate generation process into the model's internal representations. Across early-event and next-event prediction benchmarks, IVT consistently improves over text-only post-training and, under matched settings, outperforms Visual CoT on four of six splits while reducing average end-to-end inference latency by more than $5\times$. These results show that the benefits of visual reasoning do not necessarily require rendering visual thoughts explicitly.
Our analysis further shows that future-state supervision is not a universally beneficial objective. Its effectiveness depends critically on how predictive learning is integrated into the model, including the target representation, decoder architecture, prediction horizon, balance between understanding and prediction data, predictive objective, and training curriculum. These findings suggest that the key role of visual generation for reasoning may lie less in producing visually realistic intermediate images and more in providing structured predictive supervision that encourages the model to represent how the visual world may evolve.
More broadly, IVT points toward a different way of designing multimodal reasoning systems. Rather than externalizing every intermediate computation into language or pixels, models can acquire rich predictive structure during training and internalize that structure for efficient inference. For proactive reasoning, where both foresight and latency are essential, this work provides a promising path toward multimodal systems that are simultaneously predictive, grounded, and efficient.

\ifdefined\ivtanon\else
\section*{Acknowledgment}
We sincerely thank Mingfei Gao, Anshul Shah, Jiqi Yang, and Rosie Zhao for the helpful discussions.
\fi

%% file: sections/appendix.tex
\appendix
\clearpage
\renewcommand{\added}[1]{#1}

\section{Training Algorithms}
\label{app:algos}

Algorithms~\ref{alg:ivt-train} and~\ref{alg:two-stage} detail the training procedures for IVT and the two-stage baseline, respectively. For IVT, the textual and predictive objectives are computed within the same forward pass, with disjoint supervision masks identifying answer and future-prediction positions. The data-mixture ratio controls the relative frequency of understanding and prediction examples in the packed training data. Predictive targets are detached through $\operatorname{sg}(\cdot)$, so gradients from $\mathcal{L}_{\mathrm{pred}}$ do not propagate through the target branch. Frozen target encoders, including Flux-VAE, remain fixed throughout training.

\begin{algorithm}[H]
\caption{Joint training for Internalized Visual Thinking}
\label{alg:ivt-train}
\begin{algorithmic}[1]
\Require Understanding data $\mathcal{D}_{\mathrm{text}}$, prediction data $\mathcal{D}_{\mathrm{pred}}$, model parameters $\theta$, target encoder $E_{\mathrm{tar}}$, predictive-loss weight $\lambda_{\mathrm{pred}}=1$, understanding-to-prediction mixture $1{:}1$
\For{training step $s=1,\ldots,S$}
    \State Construct a packed batch according to the prescribed data mixture
    \State Encode the observed frames and prompts into an interleaved multimodal sequence
    \State Perform a single forward pass to obtain answer logits and predictive hidden states
    \State Compute the token-normalized text loss $\mathcal{L}_{\mathrm{text}}$
    \If{the batch contains future-prediction positions}
        \State Compute detached targets $Z \gets \operatorname{sg}(E_{\mathrm{tar}}(I_f))$
        \State Compute $\mathcal{L}_{\mathrm{pred}}$ using feature regression or rectified-flow matching
    \Else
        \State $\mathcal{L}_{\mathrm{pred}} \gets 0$
    \EndIf
    \State $\mathcal{L} \gets \mathcal{L}_{\mathrm{text}} + \lambda_{\mathrm{pred}}\mathcal{L}_{\mathrm{pred}}$
    \State Backpropagate $\mathcal{L}$ and update $\theta$ using AdamW
    \State Update the exponential-moving-average (EMA) weights
\EndFor
\State \Return EMA model parameters
\end{algorithmic}
\end{algorithm}

The two stages use independent optimizer states. Stage~2 initializes the model from the EMA weights obtained in Stage~1 but uses a fresh AdamW optimizer and a newly initialized learning-rate schedule.

\begin{algorithm}[H]
\caption{Two-stage predictive training followed by task fine-tuning}
\label{alg:two-stage}
\begin{algorithmic}[1]
\Require Prediction data $\mathcal{D}_{\mathrm{pred}}$, understanding data $\mathcal{D}_{\mathrm{text}}$
\State Initialize from the same BAGEL-7B-MoT checkpoint used by the other methods
\For{$s=1,\ldots,S_{\mathrm{pred}}=10\text{k}$}
    \State Optimize only the predictive objective $\mathcal{L}_{\mathrm{pred}}$
\EndFor
\State Initialize Stage~2 from the Stage-1 EMA weights
\State Reinitialize AdamW and reset the learning-rate schedule
\For{$s=1,\ldots,S_{\mathrm{text}}=20\text{k}$}
    \State Optimize only the textual objective $\mathcal{L}_{\mathrm{text}}$
\EndFor
\State \Return final model
\end{algorithmic}
\end{algorithm}

\section{Dataset Construction and Preprocessing}
\label{app:data}

For each task family, we jointly train on the native training splits of Ego-Exo4D, Ego4D, and EPIC-KITCHENS-100. We inherit the official train/validation partitions of each dataset. The three training datasets are mutually disjoint by construction. Algorithm~\ref{alg:data} summarizes the construction of proactive video reasoning examples.

\begin{algorithm}[H]
\caption{Constructing proactive video reasoning examples}
\label{alg:data}
\begin{algorithmic}[1]
\Require Annotated video $V$, task type $r$, observation ratios $\mathcal{R}=\{0.1,0.3,0.5,0.7\}$, anticipation gap $\tau_a=1$ s, prediction offsets $\mathcal{H}$
\Ensure Observed video prefix $X_{\leq t}$, textual target $y$, future-frame targets $\{I_{t+h}\}_{h\in\mathcal{H}}$
\If{$r=\textsc{EarlyEvent}$}
    \State Select an action segment $(t_s,t_e,a)$
    \For{each $\rho\in\mathcal{R}$}
        \State Set observation boundary $t \gets t_s+\rho(t_e-t_s)$
        \State Sample $X_{\leq t}$ through $t$ at $1$ fps, using at most $16$ frames
        \State Set $y$ to the native action description of $a$
        \State Extract future-frame targets $\{I_{t+h}\}_{h\in\mathcal{H}}$
    \EndFor
\ElsIf{$r=\textsc{NextEvent}$}
    \State Select context ending at $t$ and identify the next action $a_{\mathrm{next}}$ with onset $t_{\mathrm{next}}$
    \State Enforce $t \leq t_{\mathrm{next}}-\tau_a$
    \State Sample $X_{\leq t}$ through $t$ at $1$ fps
    \State Set $y$ to the native description of $a_{\mathrm{next}}$
    \State Extract future-frame targets $\{I_{t+h}\}_{h\in\mathcal{H}}$
\EndIf
\State \Return $X_{\leq t}, y, \{I_{t+h}\}_{h\in\mathcal{H}}$
\end{algorithmic}
\end{algorithm}

\section{Implementation Details}
\label{app:impl}

Table~\ref{tab:app-hparams} summarizes the training configuration used for the target-representation ablations and the main IVT experiments.

\begin{table}[H]
\centering
\small
\caption{\textbf{Training hyperparameters.} Default configuration used for the target-representation ablations and main IVT experiments. Architecture and curriculum variants modify only the corresponding design choices described in the main paper.}
\label{tab:app-hparams}
\begin{tabular}{@{}l p{0.60\linewidth}@{}}
\toprule
Configuration & Value \\
\midrule
Input resolution
& $224$--$448$ px \\

Optimizer
& AdamW ($\beta_1{=}0.9$, $\beta_2{=}0.95$, $\epsilon{=}10^{-15}$, weight decay $0$) \\

Peak learning rate
& $2\times10^{-5}$ \\

Warm-up steps
& $2{,}000$ \\

Learning-rate schedule
& Constant after warm-up \\

Gradient clipping
& $1.0$ \\

Packed tokens / step
& $36{,}864$ (VAE); $73{,}728$ (DINOv2 / SigLIP2) \\

Gradient accumulation
& None \\

Precision
& bfloat16 (autocast) \\

Predictive-loss weight $\lambda_{\mathrm{pred}}$
& $1$ ($=\texttt{mse\_weight}=\texttt{ce\_weight}$) \\

EMA decay
& $0.9999$ \\
\bottomrule
\end{tabular}
\end{table}

\section{Additional Experimental Results}
\label{app:morersts}

\subsection{Out-of-Domain Transfer to Charades}
\label{app:transfer}

\begin{table}[t]
\centering
\begin{minipage}[t]{0.44\linewidth}
\centering
\caption{\textbf{Out-of-domain transfer from Ego-Exo4D to Charades.}
For IVT, we keep the configurations used in the main results (Tables~\ref{tab:sota_egoexo}--\ref{tab:sota_ek100}), namely $H{=}1$ for Dense and $H{=}3$ for MoE.}
\label{tab:charades_transfer}
\setlength{\tabcolsep}{3pt}
\renewcommand{\arraystretch}{1.15}
\begin{tabular}{@{}lc@{}}
\toprule
Method & Acc. (\%) \\
\midrule
\multicolumn{2}{@{}l}{\textit{External methods}} \\
Qwen2.5-VL-7B        & 65.3 \\
Qwen3-VL-8B          & 64.6 \\
LLaVA-NeXT-Video-7B  & 69.6 \\
VideoLLaMA3-7B       & 70.1 \\
\midrule
\multicolumn{2}{@{}l}{\textit{BAGEL-7B-based methods}} \\
Answer-Only SFT & 71.7 \\
Text-CoT             & 71.1 \\
Visual-CoT           & 58.2 \\
IVT (Ours, MoE, $H{=}3$) & \underline{71.9} \\
\textbf{IVT (Ours, Dense, $H{=}1$)}
& \textbf{73.2} \\
\bottomrule
\end{tabular}
\end{minipage}
\hfill
\begin{minipage}[t]{0.50\linewidth}
\centering
\captionsetup{type=figure}
\caption{\textbf{Charades transfer across prediction horizons.}
Out-of-domain accuracy of Dense and MoE IVT across prediction horizons $H$, with Answer-Only SFT shown as a reference. Dense IVT exceeds Answer-Only SFT across all evaluated horizons and consistently outperforms the corresponding MoE variants.}
\label{fig:charades-horizon}
\includegraphics[width=\linewidth]{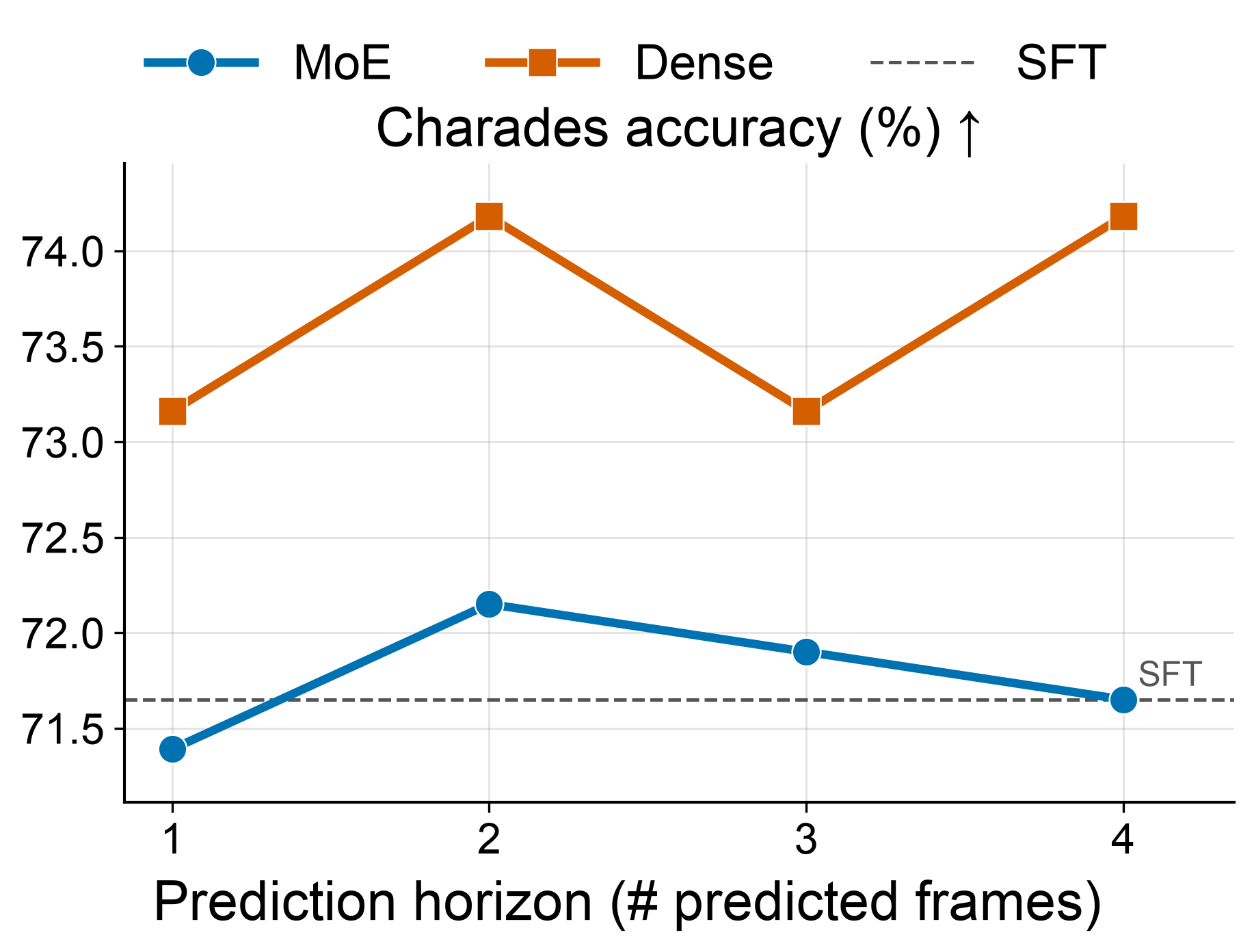}
\end{minipage}
\end{table}

The main experiments evaluate models in-domain following existing practice~\citep{sadegh2017encouraging, stergiou2023wisdom, mittal2024can}, where post-training and evaluation use the same underlying dataset distribution. We additionally examine whether representations learned through future-state prediction remain useful under distribution shift. We hold out \textbf{Charades}~\citep{sigurdsson2016hollywood} entirely from training and directly evaluate models post-trained on Ego-Exo4D early-event prediction. This setting introduces shifts in both visual content and task format. Ego-Exo4D uses open-ended descriptions of partially observed actions, whereas Charades requires selecting an answer option. Performance therefore reflects both transfer to an unseen video distribution and the preservation of instruction-following and output-format capabilities after source-domain post-training.
Table~\ref{tab:charades_transfer} shows that IVT achieves the strongest out-of-domain performance among the evaluated BAGEL-7B-based methods. Figure~\ref{fig:charades-horizon} further shows that the transfer advantage of Dense IVT persists across the prediction horizons.

\subsection{Additional Inference-Efficiency Analysis}
\label{app:eff}

\begin{algorithm}[t]
\caption{End-to-end inference-latency measurement}
\label{alg:latency}
\begin{algorithmic}[1]
\Require Model $M$, evaluation examples $\mathcal{E}$, warm-up count $W=1$
\State Fix hardware to $1\times$B200, bf16 precision, batch size $1$, decoding configuration, and maximum output length
\State Run and discard $W$ warm-up examples
\For{each example $e\in\mathcal{E}$}
    \State Load video frames from disk \Comment{excluded from timed region}
    \State \texttt{torch.cuda.synchronize()}
    \State Start wall-clock timer
    \State Execute the complete method-specific inference pipeline
    \State \texttt{torch.cuda.synchronize()}
    \State Record elapsed time $\ell_e$
\EndFor
\State \Return $\operatorname{mean}(\{\ell_e\})$ and $\operatorname{P95}(\{\ell_e\})$
\end{algorithmic}
\end{algorithm}

Our primary efficiency metric is end-to-end wall-clock latency per sample at batch size $1$, measured using the protocol in Algorithm~\ref{alg:latency}. All methods are evaluated on identical hardware and precision, with matched input-frame sampling and decoding limits. Table~\ref{tab:app-latency} reports the per-benchmark measurements underlying the aggregate results in Table~\ref{tab:vcot-prelim} and Section~\ref{sec:latency}.
Across the six dataset--task settings, Visual CoT requires an average of $6.56$ s per sample, compared with $1.20$ s for Answer-Only SFT and $1.22$ s for IVT. This corresponds to approximately $5.5\times$ and $5.4\times$ higher mean latency, respectively. The difference is also substantial in the latency tail: Visual CoT reaches a mean P95 latency of $7.92$ s, compared with $1.92$ s for Answer-Only SFT and $1.77$ s for IVT. In contrast, Answer-Only SFT and IVT exhibit closely matched latency profiles.
For Visual CoT, the timed region includes future-frame generation, VAE decoding, visual re-encoding, and final answer generation. For Answer-Only SFT and IVT, it includes input-frame encoding and answer generation. The reported measurements therefore capture the complete method-specific inference pipeline rather than an isolated model forward pass. For next-event prediction, five candidates are generated in a single deterministic beam-search call with $\texttt{num\_beams}=5$ and $\texttt{do\_sample}=\texttt{False}$, so the reported latency covers generation of the complete candidate set. Early-event prediction uses greedy decoding.

\begin{table}[H]
\centering
\small
\setlength{\tabcolsep}{3.5pt}
\renewcommand{\arraystretch}{1.12}
\caption{\textbf{End-to-end inference latency.}
Per-sample latency is measured on a single B200 GPU with batch size $1$ and bf16 precision over $200$ validation examples per benchmark. Mean and P95 latency are reported in seconds. IVT retains an inference profile comparable to Answer-Only SFT, whereas Visual CoT incurs substantially higher latency due to intermediate future-frame generation and visual re-encoding.}
\label{tab:app-latency}
\begin{tabular}{@{}ll cc cc cc@{}}
\toprule
& & \multicolumn{2}{c}{Answer-Only SFT}
& \multicolumn{2}{c}{Visual CoT}
& \multicolumn{2}{c}{IVT (Ours)} \\
\cmidrule(lr){3-4}
\cmidrule(lr){5-6}
\cmidrule(lr){7-8}
Task & Benchmark
& Mean & P95
& Mean & P95
& Mean & P95 \\
\midrule

\multirow{3}{*}{Early Event Prediction}
& Ego-Exo4D
& 1.42 & 2.45
& 7.08 & 10.54
& 1.52 & 2.34 \\

& Ego4D
& 0.83 & 1.27
& 6.38 & 7.28
& 0.81 & 1.12 \\

& EPIC-KITCHENS-100
& 0.95 & 1.64
& 6.24 & 8.34
& 0.87 & 1.30 \\

\midrule

\multirow{3}{*}{Next Event Prediction}
& Ego-Exo4D
& 1.37 & 2.40
& 7.70 & 8.62
& 1.25 & 1.91 \\

& Ego4D
& 0.92 & 1.49
& 3.72 & 4.01
& 0.82 & 0.98 \\

& EPIC-KITCHENS-100
& 1.68 & 2.26
& 8.24 & 8.74
& 2.07 & 2.96 \\

\bottomrule
\end{tabular}
\end{table}